\documentclass[journal]{IEEEtran}
\usepackage{cite}
\usepackage{amsmath,amssymb,amsfonts}
\usepackage{algorithmic}
\usepackage{graphicx}
\usepackage{textcomp}
\usepackage{xcolor}
\usepackage{hyperref}
\usepackage{multirow}
\usepackage{float}
\usepackage{amsthm}
\newtheorem{MyDef}{Definition}

\begin{document}
%
% paper title
\title{ReCoG: A Deep Learning Framework with Heterogeneous Graph for Interaction-Aware Trajectory Prediction}
% author names and IEEE memberships
\author{Xiaoyu Mo,%~\IEEEmembership{Member,~IEEE,}
        Yang Xing,~\IEEEmembership{Member,~IEEE,}
        and~Chen~Lv,~\IEEEmembership{Senior Member,~IEEE,}% <-this % stops a space
\thanks{Xiaoyu Mo and Chen Lv are with the School of Mechanical and Aerospace Engineering, Nanyang Technological University, 639798, Singapore. (e-mail: XIAOYU006@e.ntu.edu.sg, lyuchen@ntu.edu.sg)}% <-this % stops a space
\thanks{Yang Xing was with Nanyang Technological University and is now with the Department of Computer Science, University of Oxford, OX1 3QD, UK. (e-mail: yang.xing@cs.ox.ac.uk)}
}

\maketitle

%%%====================================================
\begin{abstract}
Predicting the future trajectory of surrounding vehicles is essential for the navigation of autonomous vehicles in complex real-world driving scenarios. It is challenging as a vehicle's motion is affected by many factors, including its surrounding infrastructures and vehicles. In this work, we develop the ReCoG (Recurrent Convolutional and Graph Neural Networks), which is a general scheme that represents vehicle interactions with infrastructure information as a heterogeneous graph and applies graph neural networks (GNNs) to model the high-level interactions for trajectory prediction. Nodes in the graph contain corresponding features, where a vehicle node contains its sequential feature encoded using Recurrent Neural Network (RNN), and an infrastructure node contains spatial feature encoded using Convolutional Neural Network (CNN). Then the ReCoG predicts the future trajectory of the target vehicle by jointly considering all of the features. Experiments are conducted by using the INTERACTION dataset. Experimental results show that the proposed ReCoG outperforms other state-of-the-art methods in terms of different types of displacement error, validating the feasibility and effectiveness of the developed approach.  
\end{abstract}


\IEEEpeerreviewmaketitle

%%%====================================================
\section{Introduction}
\IEEEPARstart{A}{utonomous} vehicles need to be aware of future motions of its surrounding vehicles when driving in complex and highly dynamic highway and urban roads. Trajectory prediction is promising in the context of connected vehicles, where vehicle-to-vehicle (V2V) and vehicle-to-infrastructure (V2I) wireless communications are enabled since it becomes less subjected to perception results~\cite{talebpour2016influence}.
The trajectory prediction methods for autonomous driving has progressed significantly from physics-based and maneuver-based methods to interaction-aware methods~\cite{lefevre2014survey}.  Physics-based methods~\cite{ammoun2009real} predict the future trajectory of a vehicle according to its kinematic and dynamic features. These methods ignore the effects of other vehicles and infrastructures on the target vehicle's motion so that they often fail for a prediction horizon longer than one second. Maneuver-based methods~\cite{althoff2009model} predict future motions of a target vehicle conditional on a set of maneuver prototypes. These methods take road structures into consideration for longer-term prediction but still ignore the inter-vehicular interactions. Interaction-aware methods~\cite{deo2018convolutional, mo2020interaction, gao2020vectornet}, which take driving as an interactive activity, attract more and more interest, and show better performance compared to those non-interaction-aware methods. Despite the advances, the studies of interaction-aware trajectory prediction are still limited.
Most of them consider only the interactions among vehicles ignoring traffic infrastructures. 

This work focuses on predicting the future trajectory of a target vehicle by jointly considering vehicles and infrastructures. It is assumed that vehicles' dynamics, which can be acquired by perception module or shared by connected vehicles, and the information of infrastructures, such as the local map, are available. 
Even with these assumptions, trajectory prediction is still challenging because the motion of a vehicle is affected by the driver's orientation~\cite{schwarting2019social}, interactions among vehicles and infrastructures~\cite{mo2020interaction, gao2020vectornet}. 

\begin{figure*}
    \centering
    \includegraphics[trim={0cm 0cm 0cm 0cm}, clip, width=1.0\textwidth]{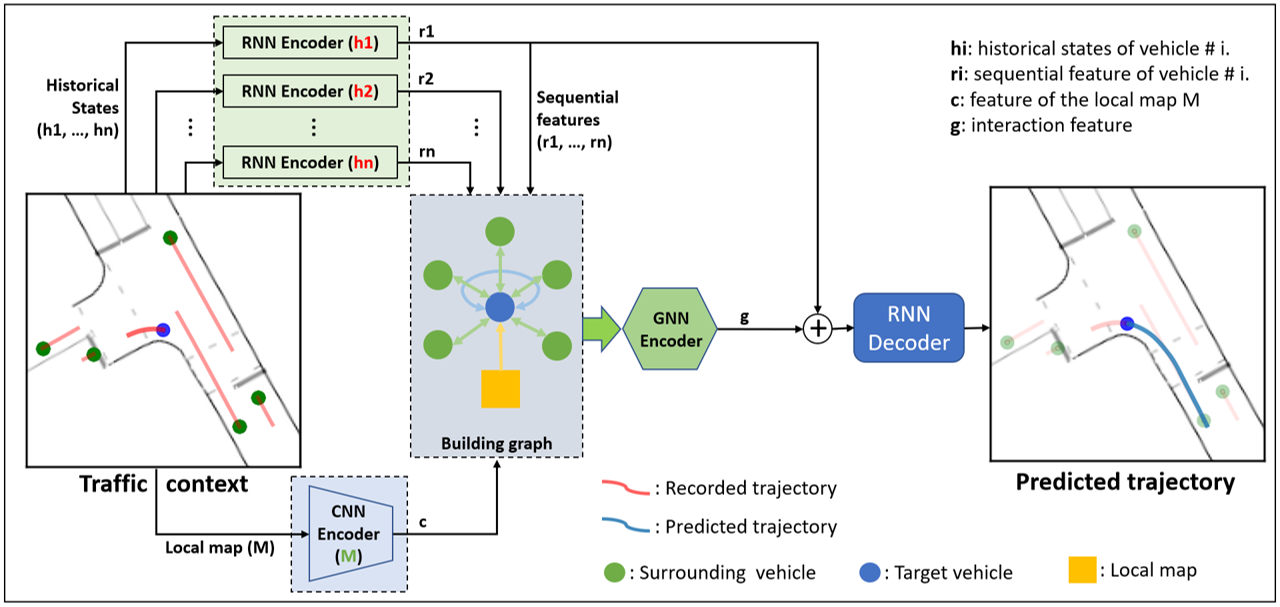}
    \vspace{-5mm}
    \caption{\textbf{The proposed scheme.} It consists of three encoders: RNN encoder for vehicle's sequential feature, CNN encoder for context feature, and GNN encoder for interaction feature. The encoded features are then sent to a RNN decoder for trajectory prediction.}
    \label{fig: girmodel}
\end{figure*}

This work proposes an interaction-aware trajectory prediction scheme that models the interaction among vehicles and infrastructures as a heterogeneous graph and then applies Graph Neural Networks (GNNs) to extract the interaction for prediction, see Fig.~\ref{fig: girmodel}. The proposed scheme adopts the Encoder-Decoder structure~\cite{cho2014learning} and uses various encoders for different inputs. It uses Recurrent Neural Networks (RNNs) to extract vehicles' dynamics from their historical states, Convolutional Neural Networks (CNNs) to recognize road structure, and GNNs to summarize the interaction from the constructed graph. The encoders constituting this scheme are decoupled to each other so that they can be replaced with more advanced encoders when available. 

This work considers the historical states consisting of positions, velocities, and yaw angles of vehicles and the simplified top-view local map. Other information can be involved if available, for instance, adding accelerations to historical states, replacing pictorial maps with High Definition (HD) Maps. 

The main contributions of this works are:
\begin{itemize}
    \item This work proposes a general scheme to model heterogeneous interactions among vehicles and infrastructures, where the encoders can be easily replaced with the latest feature extractors.
    \item This work shows that representing vehicle-infrastructure interaction as a heterogeneous graph for trajectory prediction leads to better performance than those using a homogeneous graph to represent vehicle-vehicle interactions.
    \item This work studies the performance of the proposed model with different encoders, different data availability, and different traceback horizons.
\end{itemize}

The next sections of this work are structured as follows. Sec.~\ref{sec: relatedworks} lists the related works. Sec.~\ref{sec: method} elaborates the proposed scheme. Sec.~\ref{sec: setup} sets up the experiments. Sec.~\ref{sec: results} shows the experimental results. Sec.~\ref{sec: conclusion} concludes this work and outlines future works.

%%%====================================================
\section{Related works}
\label{sec: relatedworks}
\textbf{Encoder-Decoder structures.}
The Encoder-Decoder structure was first introduced for machine translation~\cite{cho2014learning}, where one RNN encodes a variable-length sequence of symbols into a fixed-length vector, and the other RNN decodes the vector into another variable-length sequence of symbols. Auto-encoders~\cite{bengio2013generalized, doersch2016tutorial} learn to encode input data into a representation space and then decode the representation back to the input space. Auto-encoders are widely used for generation~\cite{gregor2015draw} and prediction~\cite{walker2016uncertain} tasks.
The Encoder-Decoder structure is also applicable to trajectory prediction tasks. Authors of~\cite{park2018sequence} employs LSTMs as encoder and decoder to generate future trajectories of surrounding vehicles. Convolutional social pooling~\cite{deo2018convolutional} adds convolutional layers to the LSTM encoder-decoder structure to model inter-vehicular interactions for trajectory prediction.  

\textbf{Trajectory prediction using GNNs.}
Recent interaction-aware trajectory prediction methods interpret traffic scene as a graph. Authors of~\cite{diehl2019graph} construct the graph with local connectivity, where each node contains a vehicle's feature at the current time and applies adaptions of Graph Convolutional Network (GCN)~\cite{kipf2016semi} and Graph Attention Network (GAT)~\cite{velivckovic2017graph} to model the interaction. It conceptually proves that modeling interaction as a graph improves prediction accuracy, but it considers only inter-vehicular interactions and ignores time serial features of vehicles. GRIP~\cite{li2019grip} applies graph operations on sequential features to take inter-vehicular interactions into account. It adopts the LSTM encoder-decoder structure for trajectory prediction but ignores the effects of infrastructures. SCALE-Net~\cite{jeon2020scale} applies Edge-enhanced Graph Convolutional Neural Network (EGCN) on the graph with physical states as node features to explore edge features in the constructed graph. Above mentioned methods use a homogeneous graph to model interaction, where all the nodes are vehicles only, ignoring the effects of infrastructures. Social-WaGDAT~\cite{li2020social} proposes Wasserstein Graph Double-Attention Network to model spatio-temporal interactions and employs kinematic constraints for the final prediction. It uses occupancy density maps and velocity fields generated from training data as prior of the context information. VectorNet~\cite{gao2020vectornet} uses a fully-connected hierarchical graph to model the interaction between vehicles and infrastructures, where each sub-graph represents an object and applies GNN to extract interactions. It operates on vector representations, where all the agents' trajectories, lane lines, and crosswalks are represented by vectors. The fully-connected global graph used in VectorNet is not efficient since the number of edges increases exponentially according to the number of nodes. Following VectorNet, TNT~\cite{zhao2020tnt} also represents driving context by vectors. TNT predicts multi-modal trajectories conditioned on predicted targets, then scores these trajectories and selects a subset as the final prediction. It shares the inefficient graph construction with VectorNet. EvolveGraph~\cite{li2020evolvegraph} constructs a fully-connected graph with heterogeneous agent nodes and one context node for multi-agent trajectory prediction. The same context node is linked to all target agents, while each agent should have a local context.

\textbf{Trajectory prediction using CNNs.}
CNNs are also widely explored to model interaction or road geometry for trajectory prediction. Convolutional social pooling~\cite{deo2018convolutional} defines an occupancy grid around the target vehicle, where the cell occupied by a vehicle contains that vehicle's dynamic feature encoded using LSTMs. A CNN is applied to the grid to encode interaction features without considering road structures. A decoder is then applied to the concatenation of dynamics and interaction feature to predict multi-modal trajectories conditional on predefined maneuver classes. Convolutional social pooling considers inter-vehicular interactions but ignores infrastructures. Rather than defining a spatial grid according to relative positions as in~\cite{deo2018convolutional}, authors of~\cite{mo2020interaction} use a $3\times3$ grid to contain dynamic features of the target vehicle and eight surrounding vehicles with most impacts on its driving behavior. Then the interaction feature is extracted using a two-layer CNN. Even though reducing consider vehicles, the method in~\cite{mo2020interaction} still ignores infrastructures.
Multi-agent tensor fusion (MATF)~\cite{zhao2019multi} uses CNN to encoder static scene context from a bird's eye view image then fuse the scene context with multiple agents' dynamic features to predict their future trajectories. The spatial structure is retained in MATF. It considers vehicle-infrastructure interactions, but the infrastructure is limited to a top view map. It is hard to extend MATF to consider other infrastructures in the future.

%%%====================================================
\section{Methodology}
\label{sec: method}
This section introduces the proposed scheme in detail. The problem is first formulated, followed by preliminaries on neural networks. Then the model is elaborated.
\subsection{\textbf{Problem Formulation}}
The task of this work is to predict the future trajectory of a target vehicle, based on its interaction with road and other traffic participants.

The input $\mathbf{X}_t$ to the model consists of two parts: sequences of historical states of all considered vehicles and a local map.
\begin{equation}
    \mathbf{X}_t = [\mathcal{H}_t, \mathcal{M}_t],
\end{equation}
where $\mathcal{H}_t = \lbrace h^1_t, \cdots, h^N_t \rbrace$ represents historical states of $n$ vehicles at current time $t$ with $ h^i_t = [s^i_{t-T_h+1}, s^i_{t-T_h+2}, \cdots, s^i_t]$ representing the sequence of historical states of vehicle $i$ ($i=1$ for target vehicle) at time $t$. $T_h$ is the traceback horizon. The number of considered vehicles may vary from case to case. $\mathcal{M}_t$ is the local map centered at the position of the target vehicle at time $t$.

The output is future trajectory of the target vehicle:
\begin{equation}
        f^{1}_t =  [(x^{1}_{t+1}, y^{1}_{t+1}), (x^{1}_{t+2}, y^{1}_{t+2}),\cdots, (x^{1}_{t+T_f}, y^{1}_{t+T_f})],
\end{equation}
where $T_f$ is the prediction horizon.
% \textcolor{blue}{It is worth noting that all trajectories share the same coordinate system whose origin is fixed at the position of the ego at time $t$ $(x^1_t, y^1_t)$. With this setting, trajectories of surrounding vehicles imply their relative positions with respect to the ego vehicle. this is copied from IECON}

\subsection{\textbf{Neural Networks}}
\subsubsection{\textbf{RNN}}
RNN is a class of artificial neural networks designed for tasks on data with temporal dependence, such as statistical machine translation, where the inputs and outputs are both sequences. Using two RNNs, the Encoder-Decoder~\cite{cho2014learning} method learns to encode a variable-length sequence into a vector with fixed-length, which summarizes the whole input sequence and generates a variable-length sequence by decoding the fixed-length vector. The encoder-decoder architecture is not limited to sequential data, and the encoder and decoder are not necessary to be RNNs. Long short-term memory (LSTM)~\cite{hochreiter1997long} and Gated recurrent unit (GRU)~\cite{chung2014empirical} are widely used as the cells constituting RNNs.
\subsubsection{\textbf{CNN}}
CNN is another class of artificial neural networks commonly used for tasks with image inputs, such as image classification~\cite{krizhevsky2017imagenet} and object detection~\cite{redmon2016you}. CNNs hierarchically encode image inputs into a feature vector containing the necessary information for following classification or detection tasks.
\subsubsection{\textbf{GNN}}
GNN is designed to apply neural networks to tasks with graph-like inputs, such as social network prediction and protein interface prediction~\cite{zhou2018graph, wu2020comprehensive}. Nodes in a graph contain objects, and edges between two nodes indicate their interactions. GNNs encode interaction among nodes into a feature vector by integrating information from variable-size neighborhoods.
\subsection{\textbf{Model Structure}}
The proposed method, shown in Fig.~\ref{fig: girmodel}, adopts the Encoder-Decoder structure. It is composed of three encoders (RNN encoder, CNN encoder, and GNN encoder) and one decoder (RNN decoder). The RNN encoder is applied to historical states of individual vehicles, and the CNN encoder is applied to the local map. Then the GNN encoder is used to extract the interaction feature between vehicles and the road. Finally, the encoded features are fed into the RNN decoder to predict the future trajectory of the target vehicle. These four components are elaborated on below.

\subsubsection{\textbf{RNN Encoder}}
An RNN encoder with shared weights is applied to each vehicle to capture individual sequential features from its historical states. The historical states are not restricted to XY-coordinates, which can be extended with other available information, such as velocities and yaw angles. The extendibility of this model is demonstrated in Sub.Sec.~\ref{subsec: extend}. Eq.~\ref{eq: rnn_enc} shows the RNN encoder applied to the target vehicle's historical states.
\begin{equation}
    r^1_t = {\rm FC_1}({\rm RNN_{enc}}({\rm Emb}(h^1_t))),
    \label{eq: rnn_enc}
\end{equation}
where ${\rm Emb()}$ is a linear function embedding low dimensional inputs into a higher dimensional space; ${\rm RNN_{enc}}$ is the RNN used in this work to extract sequential features from an individual vehicle's past states; ${\rm FC_1}$ is a fully connected layer applied to the RNN-encoded feature; $r^1_t$ is the sequential feature of the target vehicle. The ${\rm RNN_{enc}}$ here is not restricted to a specific form. It can be implemented using LSTM, GRU, or other state-of-the-art RNNs. The RNN encoder of the proposed model can be updated with the latest advances in the field of RNNs. The renewability of this model is demonstrated in Sub.Sec.~\ref{subsec: renew}.
\subsubsection{\textbf{CNN Encoder}}
The local map centered at the target vehicle is represented by a grayscale image and processed by a CNN. Eq.~\ref{eq: cnn_enc} shows the CNN encoder applied to the local map at time $t$.
\begin{equation}
    c_t = {\rm FC_2}({\rm CNN_{map}}(\mathcal{M}_t)),
    \label{eq: cnn_enc}
\end{equation}
where ${\rm CNN_{map}}$ is the CNN applied to the map image; ${\rm FC_2}$ is a fully connected layer applied to the CNN-encoded feature; And $c_t$ is the feature vector of the local map.
\subsubsection{\textbf{GNN Encoder}}
With encoded sequential and map features, these work models the interaction among vehicles and the road as a \textit{directed heterogeneous graph}, which contains two kinds of nodes, the \textit{vehicle node} and the \textit{map node}. 
\begin{MyDef}[Directed Heterogeneous Graph]
A graph can be represented by $\mathbb{G} = (V, E)$, where $V=\{v_1, \cdots, v_N\}$ is the set of $N$ nodes, and $E \subset V \times V$ is the set of edges.
If the edge from node $i$ to node $j$ is different from the edge from node $j$ to $node i$, the graph is a directed graph. If there are several kinds of nodes in the graph, then the graph is called a heterogeneous graph.
\end{MyDef}
To predict the target vehicle's future trajectory, both the sequential features of its neighboring vehicles and
the spatial feature of its local map are considered. A vehicle is selected as the target vehicle's neighbor if its distance to the target vehicle is within 20 meters. 
The local map is a $40\times 40 m^2$ square centered at the target vehicle.

\begin{figure}
    \centering
    \includegraphics[trim={0cm 0cm 0cm 0cm}, clip, width=0.49\textwidth]{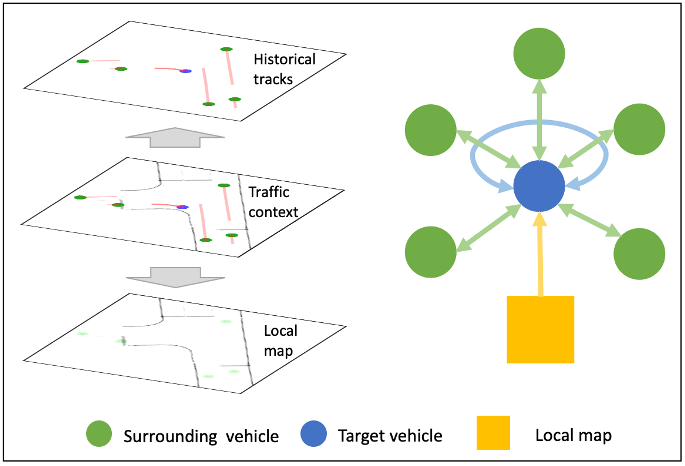}
    \vspace{-5mm}
    \caption{\textbf{Graph construction.} Left, the traffic context is split into two parts: vehicles' historical states and the local map. Right, the heterogeneous graph is constructed with local connections.}
    \label{fig: buildgraph}
\end{figure}

\textbf{Graph construction.} To model the interaction as a graph, without loss of generality, this work takes $v_1$ as the target vehicle, $v_N$ as the map node, and $\{v_2, \cdots, v_{N-1}\}$ as the neighboring vehicles. Then the edge set is determined as:
\begin{equation}
    E = \{e_{1, j}\}_{(j=1, \cdots, N-1)} \cup \{e_{j, 1}\}_{(j=1, \cdots, N)},
\end{equation}
where $e_{i,j}$ represents the directed edge from node $i$ to node $j$. An example of the constructed graph is shown in Fig.~\ref{fig: buildgraph}.

Vehicles' sequential features and map features are put into the corresponding node in the constructed, where features are concatenated with an indicator one-hot vector, [0,1] for vehicle nodes and [1,0] for road node. Then a 2-layer GNN is applied to extract high-level interaction among vehicles and the road.

\begin{equation}
    R_t = {\rm FC_1}({\rm RNN_{enc}}({\rm Emb}(\mathcal{H}_t))),
    \label{eq: rnn_ENC_ALL}
\end{equation}
\begin{equation}
    g_t = {\rm FC_3}({\rm GNN_{enc}}(R_t, c_t, E_t)),
    \label{eq: gnn}
\end{equation}
where $R_t$ is the sequential features of all vehicles, $E_t$ the edges set at time $t$, ${\rm GNN_{enc}}$ the GNN used for interaction, ${\rm FC_3}$ the fully connected layer applied to the GNN encoded feature, and $g_t$ the interaction feature.
\subsubsection{\textbf{RNN Decoder}}
Finally a RNN decoder is applied to the concatenation of the interaction feature and the target vehicle's sequential feature to predict its future trajectory.
\begin{equation}
    f^1_t = {\rm FC_4}({\rm RNN_{dec}}([g_t \Vert r^1_t ])),
    \label{eq: rnn_dec}
\end{equation}
where $[g_t \Vert r^1_t]$ is the concatenation of $g_t$ and $r^1_t$, ${\rm RNN_{dec}}$ the RNN used for prediction, and ${\rm FC_4}$ the fully connected layer mapping RNN features to a proper outputs. In this work, the output is a sequence of XY-coordinates.

%%%====================================================
\section{Experiments Setup}
\label{sec: setup}
This section describes the data (\ref{subsec: data}), metrics (\ref{subsec: metric}), and implementation details (\ref{subsec: implementation}) to set up the experiments. 

\subsection{\textbf{Data}}
\label{subsec: data}
\subsubsection{\textbf{Dataset}} 
The proposed scheme is trained and validated on the recently proposed INTERACTION dataset~\cite{interactiondataset}. It consists of various highly interactive driving situations, including highway ramps, roundabouts, and intersections, recorded using drones or fixed cameras worldwide. For each recorded scenario, it provides a high definition (HD) map, recorded vehicle tracks, and recorded pedestrian tracks. 

\begin{figure*}
    \centering
    \includegraphics[trim={0cm 0cm 0cm 0cm}, clip, width=1.0\textwidth]{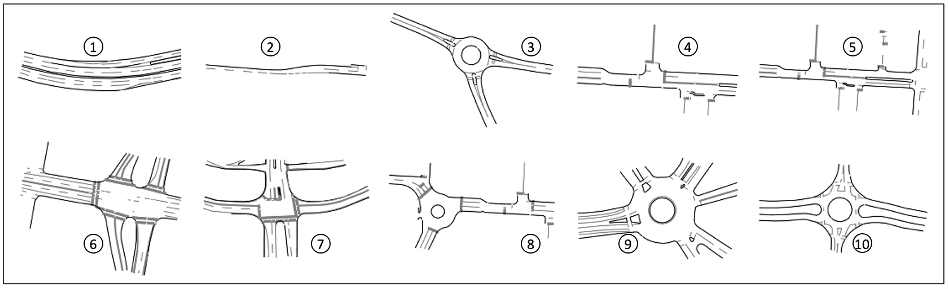}
    \vspace{-5mm}
    \caption{\textbf{Interactive driving scenarios~\cite{interactiondataset}.} The driving scenarios considered in this work including roundabout (RA, 3, 8, 9, 10), unsignalized intersection (UI, 4, 5, 6, 7), and highway ramp (HR, 1, 2).}
    \label{fig: roads10}
\end{figure*}

As shown in Fig.~\ref{fig: roads10}, ten driving scenarios, as suggested by the online benchmark~\cite{interpretchallenge}, are considered in this work. The whole dataset is split into training and validation sets, as suggested by authors of the INTERACTION dataset~\cite{interpretchallengepython}. The trajectories are split into segments of 10 seconds, where 5 seconds are saved as historical states, and the following 5 seconds as ground-truth of future trajectory. The gap between two segments is 1 second. 
After pre-processing described in the following paragraph, the training set contains 317335 data pieces and the validation set contains 90219 data pieces.

\subsubsection{\textbf{Data Pre-processing}}
This work uses a stationary frame of reference with origin fixed at the target vehicle's position at time $t$ for all trajectories. The velocities remain their recorded values while the yaw angle of the target vehicle at time $t$ is set to zero, and other yaw angle values are changed correspondingly. The local map is a $40\times 40 m^2$ square centered at the target vehicle and parallel to the target vehicle's current direction. It is represented by a $160\times 160$ array. 

\subsection{\textbf{Metrics}}
\label{subsec: metric}
This work is evaluated in terms of Displacement Error at time $\tau$ ($DE_{\tau}$), Average Displacement Error ($ADE$) over trajectories, and Final Displacement Error ($FDE$) measured in meters. The lower, the better. Many previous works reported that their work is evaluated in terms of ADE and FDE but did not provide the equation. So it is confusing about the calculation of ADE and FDE, whether it is squared error or root of squared error. This work gives the calculation of $DE_{\tau}$, $ADE$,  and $FDE$ between two trajectories as shown in Eq.~\ref{eq: DEt}, Eq.~\ref{eq: ADE}, Eq.~\ref{eq: FDE}:
\begin{equation}
    DE_{\tau} = \sqrt{(\hat{x}_{\tau} - x_{\tau})^{2} + (\hat{y}_{\tau} - y_{\tau})^{2}},
    \label{eq: DEt}
\end{equation}
\begin{equation}
    ADE = \frac{1}{T_f}\sum^{T_f}_{\tau=1} DE_{\tau} =\frac{1}{T_f}\sum^{T_f}_{t=1} \sqrt{(\hat{x}_{\tau} - x_{\tau})^{2} + (\hat{y}_{\tau} - y_{\tau})^{2}},
    \label{eq: ADE}
\end{equation}
\begin{equation}
    FDE = DE_{T_f} = \sqrt{(\hat{x}_{T_f} - x_{T_f})^{2} + (\hat{y}_{T_f} - y_{T_f})^{2}},
    \label{eq: FDE}
\end{equation}
where $(\hat{x}_{\tau}, \hat{y}_{\tau})$ is the predicted position in XY-coordinates at time $\tau$; $(x_{\tau}, y_{\tau})$ is the ground truth position at time ${\tau}$; and $T_f$ is the prediction horizon.

\subsection{\textbf{Implementation Details}}
\label{subsec: implementation}
This work is implemented using PyTorch~\cite{NEURIPS2019_9015} for the overall structure and PyTorch Geometric~\cite{Fey/Lenssen/2019} for GNN layers. The model is trained end-to-end for 10 epochs using Adam~\cite{kingma2014adam} with scheduled learning rate, which starts from 0.001 and reduces by half at the end of epochs 1,2,4, and 6, to minimize $ADE$ as defined in Eq.~\ref{eq: ADE}.
The historical states are first embedded into a 64-dimensional space then sent to the RNN encoder. The RNN encoder is a one-layer RNN with a hidden size equals to 64, and the RNN decoder is a two-layer RNN with a hidden size equals to 128. The CNN encoder is a three-layer convolutional block with $[\#\_filters, kernel\_size, stride] = [8, 16, 4]$ for the first layer, $[16, 8, 4]$ for the second layer, and $[32, 4, 2]$ for the third layer. The structure is [[Conv, LeakyReLU, BatchNorm], [Conv, LeakyReLU, BatchNorm], [Conv, LeakyReLU, BatchNorm], FC, FC]. 
The GNN encoder uses two GNN layers to extract higher-level interactions. Leaky-ReLU with a negative slope equals to 0.1 is used as the only activation function throughout the implementation. Further details can be found in the released code.
%%%====================================================
\section{Results}
\label{sec: results}
This section first shows the experiments conducted to show the extendibility (\ref{subsec: extend}) and renewability (\ref{subsec: renew}) of the proposed shceme. After that, ablative studies (\ref{subsec: ablation}) are done to show the effectiveness of the proposed model. Then the effects of traceback horizon on the prediction performance is studied (\ref{subsec: inputlength}). Finally, the proposed model is compared to the state-of-the-art methods (\ref{subsec: sota}) on the INTERACTION dataset.
\subsection{\textbf{Extendibility}}
\label{subsec: extend}
The proposed scheme models the interaction between vehicles and infrastructures as a heterogeneous graph. It is limitless to the input types. In this work, the heterogeneous graph has two types of nodes, the vehicle node, and the road node. It can be extended with other kinds of nodes to take other infrastructures, such as the traffic signals, or other agent types, such as pedestrian and bicycles, into consideration.
\begin{figure}
    \centering
    \includegraphics[trim={0cm 0cm 0cm 0cm}, clip, width=0.49\textwidth]{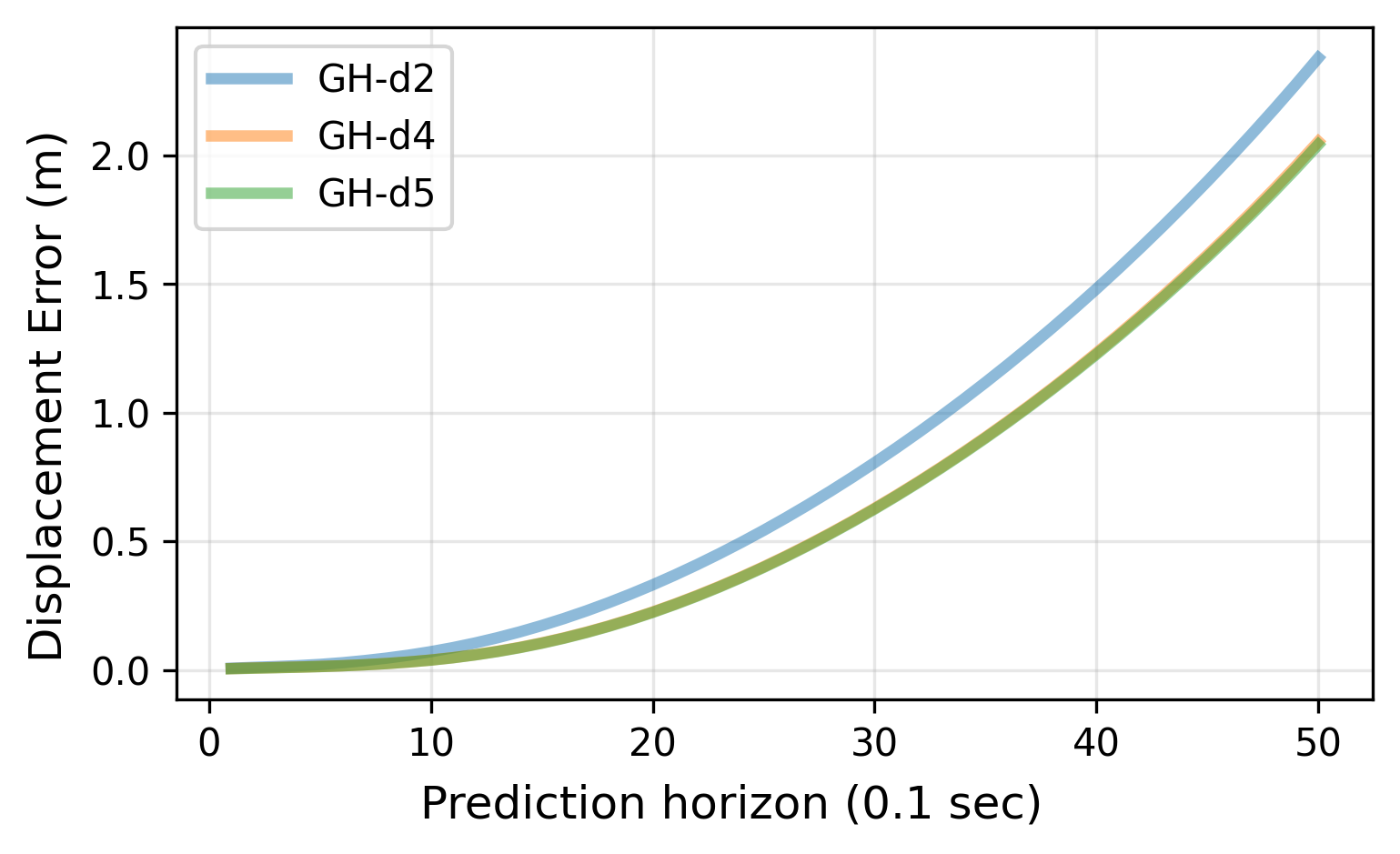}
    \vspace{-5mm}
    \caption{\textbf{Extendibility of the proposed scheme.} GH-d2: the proposed model ReCoG using $(x,y)$ as historical states; GH-d4: the the proposed model using $(x,y,v_x,v_y)$ as historical states; GH-d5: the the proposed model using $(x,y,v_x,v_y,\psi)$ as historical states. Traceback horizon: $T_h=30$ (3 sec). Prediction horizon $T_f=50$ (5 sec). GNN encoder: GAT. RNN encoder: GRU.}
    \label{fig: girhet245}
\end{figure}

For the RNN encoder, the proposed scheme does not assume the information to be considered. It can be easily extended with richer information when available. To show that, the proposed model is applied to three different inputs, $(x,y)$, $(x,y,v_x,v_y)$, and $(x,y,v_x,v_y,\psi)$, where $(x, y)$ is the position, $(v_x,v_y)$ the velocity, and $\psi$ the yaw angle of the considered vehicle.
The results are shown in Fig.~\ref{fig: girhet245}. It is shown that considering velocities in addition to positions reduces the displacement error, but further adding yaw angle shows no noticeable improvement. The scheme implemented in this section takes GAT as the GNN encoder and GRU as the RNN encoder.

This work uses CNN to extract spatial feature from a top view image of the local map. The top view image can be replaced by a vectorized representation of a HD map to reduce model size as shown in~\cite{gao2020vectornet}.
\subsection{\textbf{Renewability}}
\label{subsec: renew}
The encoders used in this work are decoupled from each other so that they can be replaced without affecting other encoders. The proposed model does not assume what recurrent unit to be used for the RNN encoder, which could be LSTM~\cite{hochreiter1997long}, GRU~\cite{chung2014empirical}, or others. This work applies a GNN to encode interactions between vehicles and infrastructures for the constructed graph with corresponding features. The GNN could be GCN~\cite{kipf2016semi}, GAT~\cite{velivckovic2017graph}, or others. 

\begin{figure}
    \centering
    \includegraphics[trim={0cm 0cm 0cm 0cm}, clip, width=0.49\textwidth]{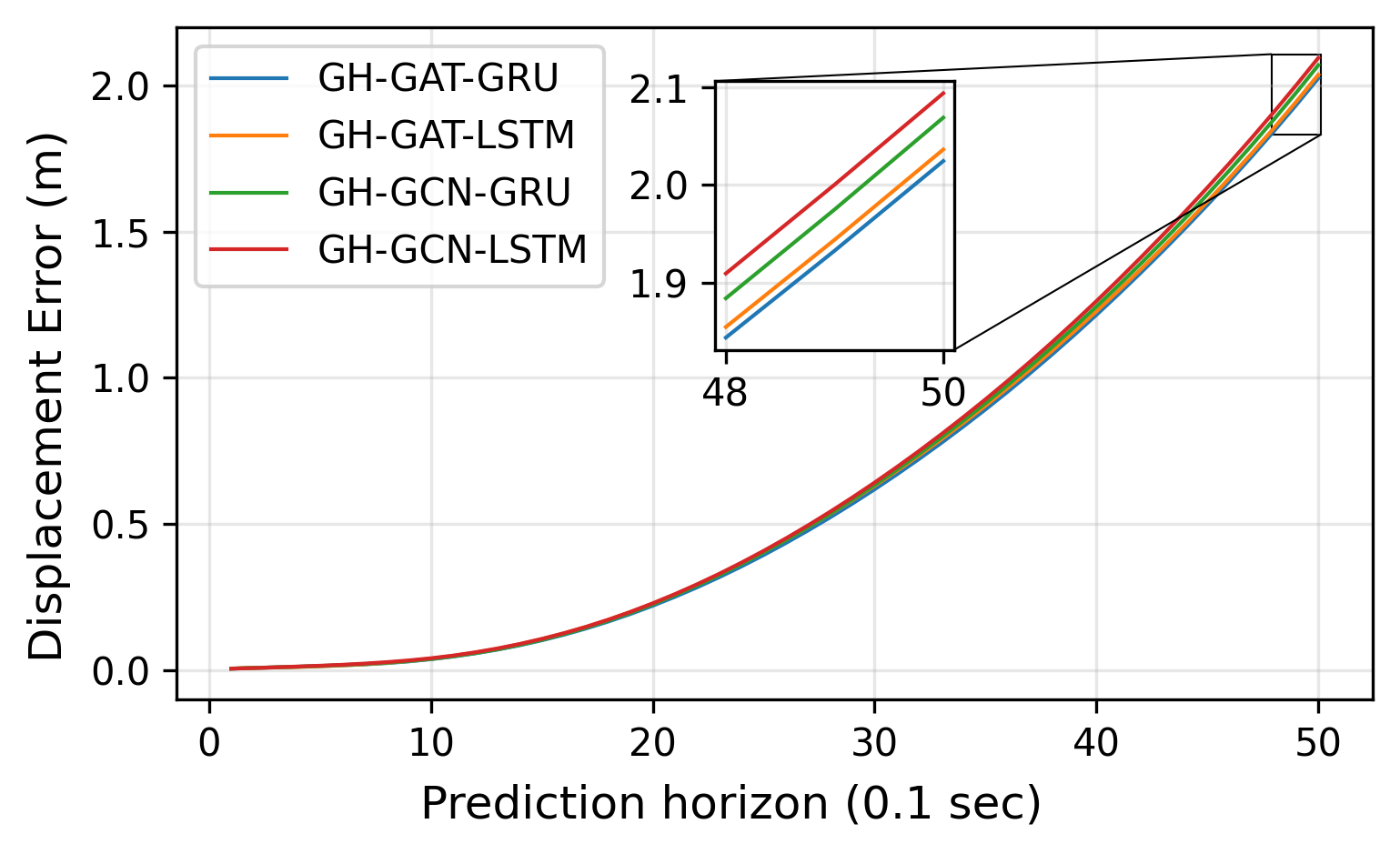}
    \vspace{-5mm}
    \caption{\textbf{Renewability of the proposed scheme.} GAT-GRU: the proposed model using GAT as GNN encoder and GRU as RNN encoders. The same rule applies to GAT-LSTM, GCN-GRU, and GCN-LSTM. Traceback horizon: $T_h=30$. Prediction horizon: $T_f=50$. Inputs data: $(x,y,v_x,v_y)$}
    \label{fig: renew}
\end{figure}

This section implements the proposed model with different RNN and GNN encoders and studies the performance of their combinations. GCN is selected because of its simplicity and effectiveness, and GAT is selected since it treats neighboring nodes differently using the attention mechanism. The results are shown in Fig.~\ref{fig: renew} and suggest that the GAT-based methods work better than the GCN-based methods, and the combination of GAT and GRU has the best performance on the trajectory prediction task studied in this work.

\subsection{\textbf{Ablative Studies}}
\label{subsec: ablation}
Ablative studies are conducted to show the advance of the proposed scheme. 

\subsubsection{R} This model predicts the future trajectory of a vehicle using only its individual sequential feature encoded by a single RNN. No interaction is considered in this model.
\subsubsection{GR} This model builds a homogeneous graph to model the interaction between vehicles, where each node represents a vehicle. It considers the interactions among vehicles but ignores the effects of infrastructures.
\subsubsection{GH} This is an implementation of the proposed scheme ReCoG, which models the interaction between vehicles and the infrastructure as a heterogeneous graph. 

\begin{figure}
    \centering
    \includegraphics[trim={0cm 0cm 0cm 0cm}, clip, width=0.49\textwidth]{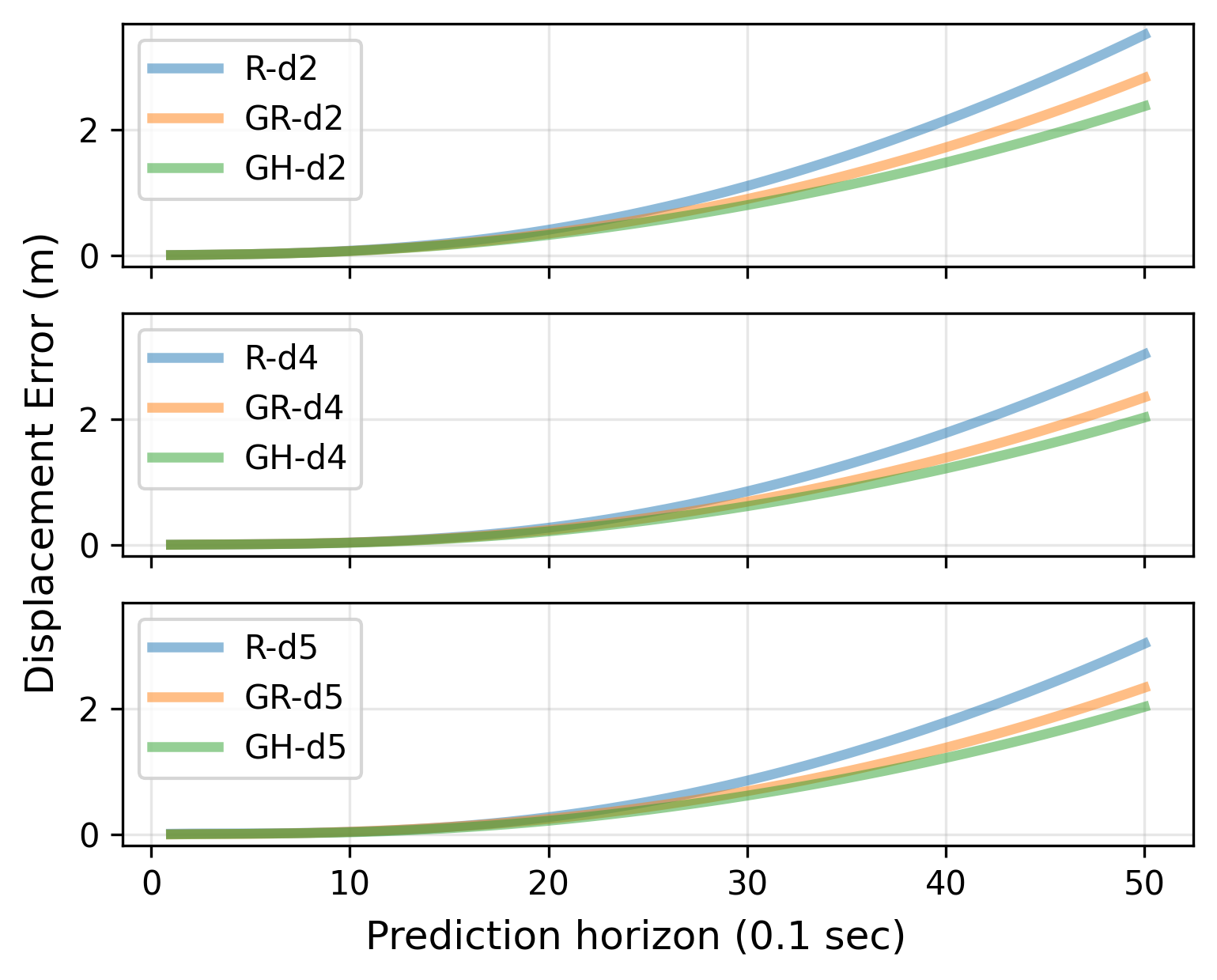}
    \vspace{-5mm}
    \caption{\textbf{Ablative studies of the proposed scheme.} Top, ablative study with $(x,y)$ as inputs; Middle, ablative study with $(x,y,v_x,v_y)$ as inputs; Bottom, ablative study with $(x,y,v_x,v_y,\psi)$ as inputs. Traceback horizon: $T_h= 30$. Prediction horizon: $T_f= 50$. GNN encoder: GAT. RNN encoder: GRU.}
    \label{fig: ablation}
\end{figure}

The ablative studies of the proposed model is conducted with different inputs. The results are shown in Fig.~\ref{fig: ablation}. It can be seen that the interaction-aware models (GR and GH) outperform the non-interaction-aware model (R) with all kinds of inputs. This result is consistent with previous works~\cite{deo2018convolutional, mo2020interaction} and demonstrates the necessity of modeling interactions for trajectory prediction. Also, the heterogeneous interaction-aware method (GH) outperforms the homogeneous interaction-aware method (GR) for all kinds of inputs. This result shows the advance of the proposed method: 1) Model vehicle-infrastructure interactions as a heterogeneous graph; 2) Obtain features of different nodes with different encoders; 3) Use GNN to extract the interaction feature. This work considers two kinds of nodes for demonstration, while other kinds of nodes can be further considered under the proposed framework.

\subsection{\textbf{Effects of Traceback Horizon}}
\label{subsec: inputlength}
\begin{figure}
    \centering
    \includegraphics[trim={0cm 0cm 0cm 0cm}, clip, width=0.49\textwidth]{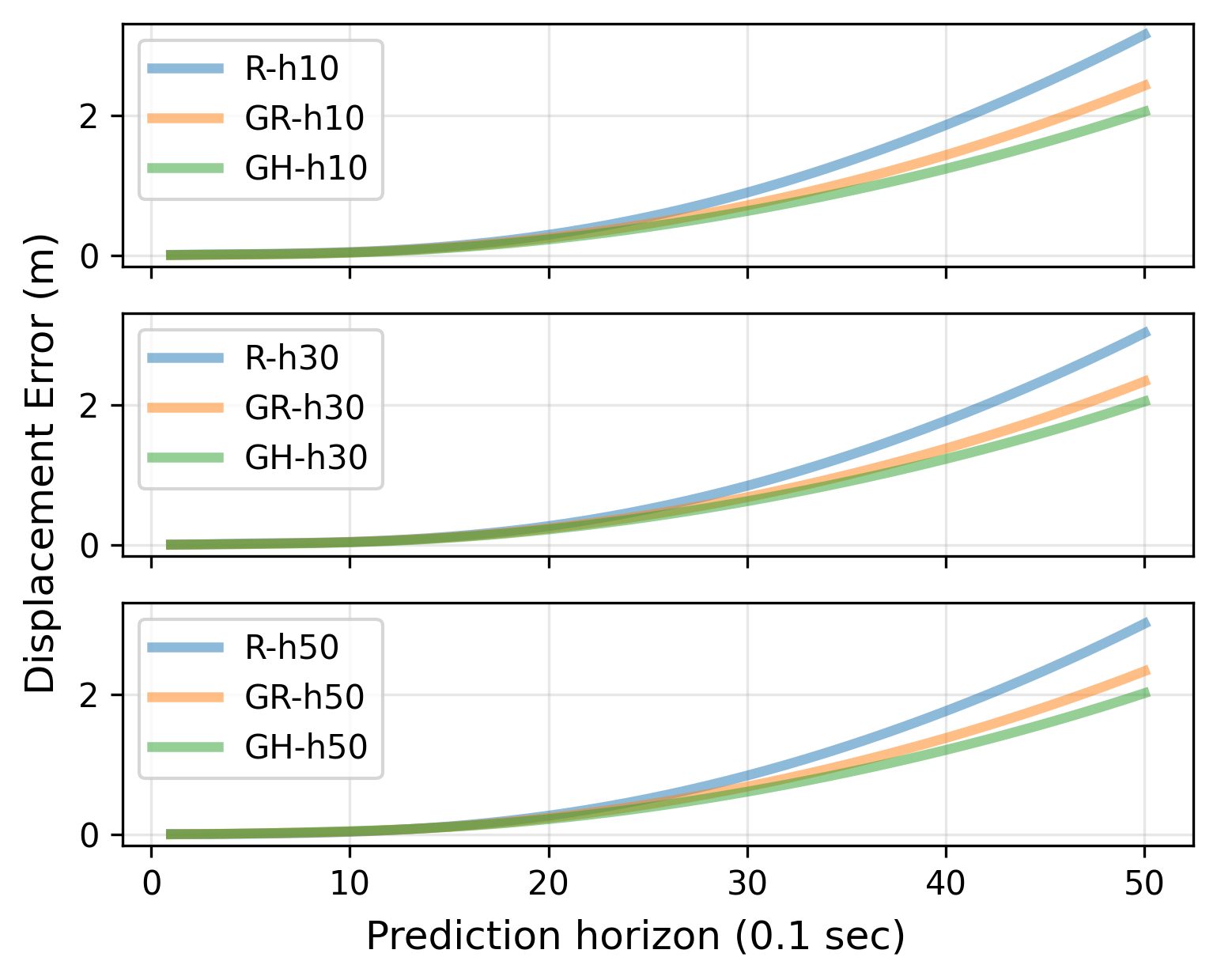}
    \vspace{-5mm}
    \caption{\textbf{Effects of the traceback horizon.} Traceback horizon: $T_h= 10, 30, 50$, respectively. Prediction horizon: $T_f= 50$. Inputs data: $(x,y,v_x,v_y)$, GNN encoder: GAT. RNN encoder: GRU.}
    \label{fig: inputlength}
\end{figure}

This section studies how the traceback horizon affects the prediction accuracy of the proposed scheme and its ablations. The traceback horizons ($T_h$) are set to three values ($10, 30, 50$) and studied with the input data $(x,y,v_x,v_y)$. Fig.~\ref{fig: inputlength} shows that prolonging $T_h$ from $10$ to $30$ leads to better performance for R and GR while further prolonging $T_h$ to $50$ shows very limited improvement.
For GH, the differences among different traceback horizons are slight. To keep the traceback horizon consistent, this work conducts other experiments with $T_h=30$. Fig.~\ref{fig: inputlength} also shows that the effects of traceback horizons varies for different methods. 

\subsection{\textbf{Comparison With State-of-the-art}}
\label{subsec: sota}
The proposed scheme is trained and validated with the dataset used in other sections to compare with state-of-the-art methods. But the traceback horizon and prediction horizon are set to $T_h=10$ and $T_f=30$, respectively. The results are further submitted to the INTERPRET challenge leader-board~\cite{interpretchallenge}. 

Tab.~\ref{tab: othersresults} compares the proposed scheme and its ablations to state-of-the-art methods on the INTERACTION validation set and shows the results on the online benchmark. The results of other methods in Tab.~\ref{tab: othersresults} are reported in~\cite{zhao2020tnt}. It is shown that ReCoG outperforms existing methods and achieves the state-of-the-art performance on the INTERACTION validation set. But results on the online benchmark (Test set) shows that ReCoG is more overfitted to the the training set comparing to its ablation, the GR model. 

The overfitting problem of ReCoG is solved with enlarged training set, where the trajectories are split into segments of 4 seconds and the gap between two segments are set to 0.1 second,  (Sub.Sec.~\ref{subsec: data}). The proposed scheme is implemented with larger size of parameters and trained with enlarged dataset. Results on the online benchmark shows that ReCoG (named GH for submission) outperforms all its competitors, including its ablation GR. Finally, ReCoG (named GH\_29\_3 for submission) won the INTERPRET Challenge~\cite{interpretchallenge}, which belongs to the Thirty-fourth Conference on Neural Information Processing Systems (NeurIPS 2020) competition track~\cite{neurips2020competition}, with its fine-tuned versions. The achieved ADE is $\textbf{0.1878}$ meter and the FDE is $\textbf{0.6381}$ meter.

\subsection{\textbf{Visualized results}}
\begin{figure*}
    \centering
    \includegraphics[trim={0cm 0cm 0cm 0cm}, clip, width=1.0\textwidth]{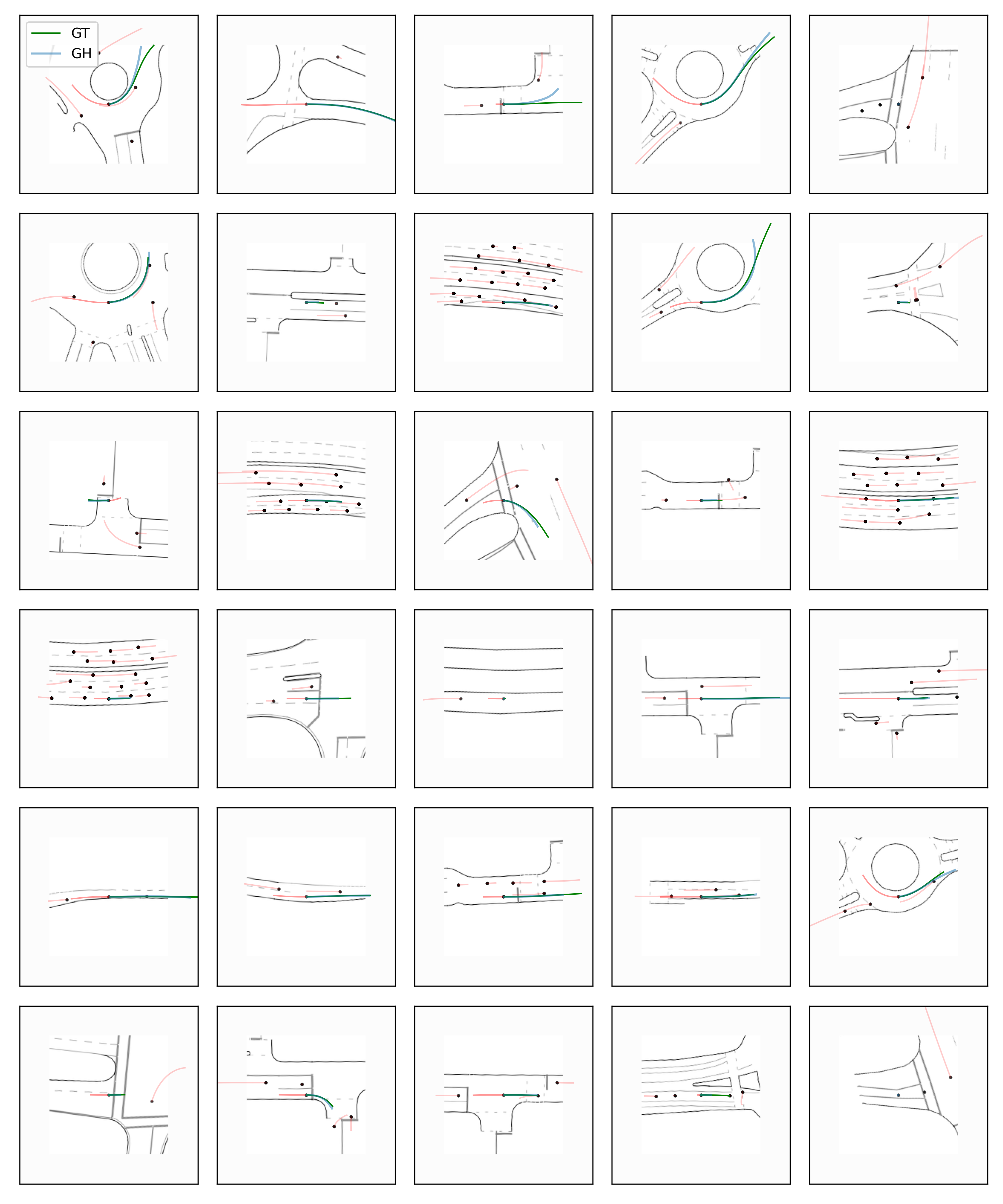}
    \caption{\textbf{Qualitative results of ReCoG on the INTERACTION validation dataset.} Traceback horizon: $T_h= 30$. Prediction horizon: $T_f= 50$. Inputs data: $(x,y,v_x,v_y)$, GNN encoder: GAT. RNN encoder: GRU. GT: Ground-Truth trajectory. GH: ReCoG predicted trajectory }
    \label{fig: visualization}
\end{figure*}

Visualization of the ReCoG predictions on the INTERACTION validation set is shown in Fig.~\ref{fig: visualization}, where GT is the ground-truth future trajectory and GH is the corresponding predicted trajectory. Fig.~\ref{fig: visualization} shows that the proposed scheme ReCoG is able to predict future trajectory of the target vehicle under various driving situations.

\begin{table}
\centering
\caption{\textbf{Comparison with state-of-the-art methods on INTERACTION dataset}}
\centering
\begin{tabular}{|c|c c|c c|} 
\hline
\multirow{2}{6em}{ \textbf{Methods }} & \multicolumn{2}{c|}{\textbf{Validation set ($T_f=30$)}} & \multicolumn{2}{c|}{\textbf{Test set ($T_f=30$)}}\\
\cline{2-5} 
& ADE (m) & FDE (m) & ADE (m) & FDE (m)\\
\hline
DESIRE~\cite{lee2017desire} & 0.32 & 0.88 & - &  -\\
MultiPath~\cite{chai2019multipath} & 0.30 & 0.99 & - & -\\
TNT~\cite{zhao2020tnt} & 0.21 & 0.67 & - & -\\
\hline
\hline
R & 0.2527 & 0.9000 & 0.4508 & 1.4702\\
GR & 0.2098 & 0.7202 & \textbf{0.3559} & \textbf{1.1715}\\
ReCoG & \textbf{0.1919} & \textbf{0.6462} & 0.3940 & 1.2555\\
\hline
\end{tabular}
\vspace{0.2cm}
\label{tab: othersresults}
\end{table}

% Another tabel showing TDE for 10 scenarios. Or three kinds of scenarios as in~\cite{li2020social}
%%%====================================================
\section{Conclusion}
\label{sec: conclusion}
This study proposes a general scheme to model vehicle-infrastructure interactions for vehicular trajectory prediction. The proposed scheme constructs a heterogeneous graph to represent the interactions, where the nodes contain features from corresponding encoders, then applies graph neural networks to extract interaction features. Experimental results on the recently released dataset show that modeling vehicle-infrastructure interaction improves accuracy comparing to the method considering inter-vehicular interaction only, and the proposed scheme matches state-of-the-art methods. The proposed scheme consists of decoupled encoders and a decoder, which can be easily updated with the most advanced feature extractors.

One way to improve the proposed scheme is to consider a more-informative HD map rather than a pictorial one. This scheme is not limited to two node types, and other types of nodes, such as traffic signals, can be easily involved with proper encoders. The proposed scheme predicts the future trajectory for just one target vehicle, which should be generalized to predict trajectories for multiple surrounding vehicles to support the downstream decision-making and planning modules. Other graph construction strategies can also be investigated to see how they affect performance. 

\appendices
\section{Detailed Results}
Tab.~\ref{tab: detailedresults} shows the detailed ADE and FDE values of the implemented models (Method), trained and validated with dataset described in Sub.Sec.~\ref{subsec: data}, on different driving scenarios (Scene), where scenarios are numbered consistent with Fig.~\ref{fig: roads10}. It can be seen that, RoCoG outperforms its ablations throughout all the scenarios.

\begin{table*}
\centering
\caption{\textbf{Detailed quantitative results in ADE / FDE $(m)$ for $T_h=30$ and $s_t=(x,y,v_x,v_y)$}.}
\centering
\begin{tabular}{|c|c|c|c|c|c|c|} 
\hline
\multirow{2}{*}{\textbf{Scene}} & \multirow{2}{*}{\textbf{Method}} & \multicolumn{5}{c|}{\textbf{Prediction horizon}}\\
\cline{3-7}
 &  & 1.0s & 2.0s & 3.0s & 4.0s & 5.0s\\
\hline
\hline
\multirow{3}{*}{\textbf{1}} & R & 0.0121 / 0.0276 & 0.0589 / 0.2034 & 0.1763 / 0.615 & 0.366 / 1.2225 & 0.6183 / 1.9784\\
& GR & 0.0115 / 0.0263 & 0.0487 / 0.1568 & 0.1288 / 0.4121 & 0.2447 / 0.7527 & 0.3926 / 1.1912\\
& GH & \textbf{0.0101} / \textbf{0.023} & \textbf{0.0451} / \textbf{0.1481} & \textbf{0.1222} / \textbf{0.3958} & \textbf{0.2345} / \textbf{0.7279} & \textbf{0.3788} / \textbf{1.1578}\\\hline
\hline
\multirow{3}{*}{\textbf{2}} & R & 0.0195 / 0.0437 & 0.0863 / 0.2879 & 0.2507 / 0.8727 & 0.5301 / 1.8304 & 0.9304 / 3.1607\\
& GR & 0.02 / 0.0428 & 0.0761 / 0.2362 & 0.1977 / 0.64 & 0.3887 / 1.2578 & 0.6527 / 2.117\\
& GH & \textbf{0.0185} / \textbf{0.0378} & \textbf{0.0678} / \textbf{0.2089} & \textbf{0.1724} / \textbf{0.5467} & \textbf{0.3297} / \textbf{1.0358} & \textbf{0.5422} / \textbf{1.7086}\\\hline
\hline
\multirow{3}{*}{\textbf{3}} & R & 0.0429 / 0.0938 & 0.1681 / 0.536 & 0.4662 / 1.5894 & 0.9801 / 3.4339 & 1.7768 / 6.3827\\
& GR & \textbf{0.036} / 0.0758 & 0.147 / 0.4929 & 0.4319 / 1.5193 & 0.9331 / 3.3296 & 1.7024 / 6.1131\\
& GH & 0.0428 / \textbf{0.0816} & \textbf{0.1298} / \textbf{0.3783} & \textbf{0.3256} / \textbf{1.0632} & \textbf{0.6667} / \textbf{2.297} & \textbf{1.2017} / \textbf{4.3426}\\\hline
\hline
\multirow{3}{*}{\textbf{4}} & R & 0.0254 / 0.0542 & 0.0994 / 0.3201 & 0.2789 / 0.9565 & 0.579 / 1.9574 & 0.9927 / 3.2482\\
& GR & 0.0253 / 0.0545 & 0.0941 / 0.2946 & 0.2519 / 0.8372 & 0.508 / 1.6797 & 0.8618 / 2.8047\\
& GH & \textbf{0.0234} / \textbf{0.0502} & \textbf{0.0865} / \textbf{0.2685} & \textbf{0.2269} / \textbf{0.7403} & \textbf{0.4465} / \textbf{1.4326} & \textbf{0.7374} / \textbf{2.3117}\\\hline
\hline
\multirow{3}{*}{\textbf{5}} & R & 0.0275 / 0.0597 & 0.1069 / 0.3386 & 0.2922 / 0.9885 & 0.6025 / 2.0382 & 1.0393 / 3.4463\\
& GR & \textbf{0.027} / 0.0582 & 0.1017 / 0.3179 & 0.2719 / 0.9053 & 0.5509 / 1.8362 & 0.9416 / 3.0947\\
& GH & \textbf{0.027} / \textbf{0.0566} & \textbf{0.0971} / \textbf{0.2972} & \textbf{0.2515} / \textbf{0.8184} & \textbf{0.4959} / \textbf{1.6023} & \textbf{0.8264} / \textbf{2.6341}\\\hline
\hline
\multirow{3}{*}{\textbf{6}} & R & 0.0212 / 0.0473 & 0.09 / 0.2975 & 0.2626 / 0.9244 & 0.564 / 1.9771 & 1.0026 / 3.463\\
& GR & 0.0218 / 0.0471 & 0.086 / 0.2771 & 0.2415 / 0.8286 & 0.506 / 1.7409 & 0.8881 / 3.0291\\
& GH & \textbf{0.0203} / \textbf{0.0441} & \textbf{0.0803} / \textbf{0.2579} & \textbf{0.223} / \textbf{0.7585} & \textbf{0.4604} / \textbf{1.5564} & \textbf{0.7963} / \textbf{2.666}\\
\hline
\hline
\multirow{3}{*}{\textbf{7}} & R & 0.024 / 0.0559 & 0.1159 / 0.4038 & 0.366 / 1.3454 & 0.819 / 2.9557 & 1.4837 / 5.2434\\
& GR & 0.0259 / 0.0581 & 0.1114 / 0.3704 & 0.3263 / 1.147 & 0.6987 / 2.441 & 1.2403 / 4.3134\\
& GH & \textbf{0.0235} / \textbf{0.052} & \textbf{0.0982} / \textbf{0.3234} & \textbf{0.2818} / \textbf{0.974} & \textbf{0.5891} / \textbf{2.0049} & \textbf{1.0212} / \textbf{3.4444}\\
\hline
\hline
\multirow{3}{*}{\textbf{8}} & R & 0.0251 / 0.057 & 0.1132 / 0.3837 & 0.3396 / 1.2063 & 0.7299 / 2.5401 & 1.2775 / 4.2798\\
& GR & 0.025 / 0.055 & 0.1022 / 0.3352 & 0.2924 / 1.0126 & 0.6144 / 2.1038 & 1.0698 / 3.5969\\
& GH & \textbf{0.0237} / \textbf{0.0506} & \textbf{0.0927} / \textbf{0.2972} & \textbf{0.2534} / \textbf{0.8447} & \textbf{0.5075} / \textbf{1.655} & \textbf{0.8503} / \textbf{2.7286}\\
\hline
\hline
\multirow{3}{*}{\textbf{9}} & R & 0.025 / 0.0545 & 0.1088 / 0.3703 & 0.3322 / 1.1984 & 0.7311 / 2.613 & 1.318 / 4.6179\\
& GR & 0.0264 / 0.0557 & 0.1047 / 0.3438 & 0.304 / 1.0683 & 0.6516 / 2.2798 & 1.1558 / 3.9752\\
& GH & \textbf{0.0238} / \textbf{0.0512} & \textbf{0.094} / \textbf{0.3052} & \textbf{0.2667} / \textbf{0.92} & \textbf{0.5582} / \textbf{1.907} & \textbf{0.9704} / \textbf{3.2597}\\
\hline
\hline
\multirow{3}{*}{\textbf{10}} & R & 0.0246 / 0.0526 & 0.0947 / 0.2984 & 0.2567 / 0.864 & 0.5267 / 1.7711 & 0.9022 / 2.9615\\
& GR & 0.0249 / 0.0531 & 0.0886 / 0.2683 & 0.2257 / 0.7256 & 0.4426 / 1.4318 & 0.7405 / 2.3781\\
& GH & \textbf{0.0223} / \textbf{0.0473} & \textbf{0.0795} / \textbf{0.242} & \textbf{0.2021} / \textbf{0.6449} & \textbf{0.3908} / \textbf{1.2415} & \textbf{0.6454} / \textbf{2.0413}\\
\hline
\end{tabular}
\vspace{0.2cm}
\label{tab: detailedresults}
\end{table*}

% \section*{Acknowledgment}
% The authors would like to thank...

% Can use something like this to put references on a page
% by themselves when using endfloat and the captionsoff option.
\ifCLASSOPTIONcaptionsoff
  \newpage
\fi

\bibliographystyle{IEEEtran}
\bibliography{GIRhet.bib}

\begin{IEEEbiography}[{\includegraphics[width=1in,height=1.25in,clip,keepaspectratio]{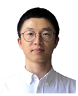}}]{Xiaoyu Mo} received the B.E. degree from Yangzhou University, Yangzhou, China, in 2015, the M.E. degree from Huazhong University of Science and Technology, Wuhan, China, in 2017. He was a research associate with Nanyang Technological University, Singapore,  from Sep. 2017 to Jan. 2019. He is now a Ph.D. candidate at Nanyang Technological University. His research interests include trajectory prediction and decision making for connected autonomous vehicles.
\end{IEEEbiography}

\begin{IEEEbiography}[{\includegraphics[width=1in,height=1.25in,clip,keepaspectratio]{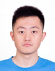}}]{Yang Xing} received his Ph. D. degree from Cranfield University, UK, in 2018. He is currently a Research Associate with the Department of Computer Science, University of Oxford, UK. He was a Research Fellow with the School Mechanical and Aerospace Engineering, Nanyang Technological University, Singapore.  His research interests include machine learning, human behavior modeling, intelligent multi-agent collaboration, and intelligent/autonomous vehicles. His work focuses on the understanding of human behaviors using machine learning methods, and intelligent and automated vehicle design. He received the IV2018 Best Workshop/Special Issue Paper Award. Dr. Xing serves as a Guest Editor for IEEE Internet of Thing, IEEE Intelligent Transportation Systems Magazine, and Frontiers in Mechanical Engineering. He is also an active reviewer for IEEE Transactions on Intelligent Transportation Systems, Vehicular Technology, Industrial Electronics, and IEEE/ASME Transactions on Mechatronics, etc. 
\end{IEEEbiography}

\begin{IEEEbiography}[{\includegraphics[width=1in,height=1.25in,clip,keepaspectratio]{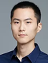}}]{Chen Lv} is currently an Assistant Professor at School of Mechanical and Aerospace Engineering, and the Cluster Director in Future Mobility Solutions at ERI@N, Nanyang Technology University, Singapore. He received the Ph.D. degree at the Department of Automotive Engineering, Tsinghua University, China in 2016. He was a joint PhD researcher at EECS Dept., University of California, Berkeley, USA during 2014-2015, and worked as a Research Fellow at Advanced Vehicle Engineering Center, Cranfield University, UK during 2016-2018. His research focuses on advanced vehicles and human-machine systems, where he has contributed over 100 papers and obtained 12 granted patents. 
\end{IEEEbiography}

\end{document}